\documentclass{article}
\pdfoutput=1

\PassOptionsToPackage{numbers, super}{natbib}
\usepackage[preprint]{neurips_2019}

\usepackage[title]{appendix}
\usepackage[utf8]{inputenc} % allow utf-8 input
\usepackage[T1]{fontenc}    % use 8-bit T1 fonts
\usepackage{hyperref}       % hyperlinks
\usepackage{url}            % simple URL typesetting
\usepackage{booktabs}       % professional-quality tables
\usepackage{amsmath}
\usepackage{amssymb}
\usepackage{amsfonts, bm}       % blackboard math symbols
\usepackage{nicefrac}       % compact symbols for 1/2, etc.
\usepackage{mathtools}
\usepackage{microtype}      % microtypography
\usepackage{graphicx}
\usepackage{enumitem}
\usepackage{subfig}
\usepackage{color}
\usepackage[table]{xcolor}  % for coloring table
\newcolumntype{L}{>{\centering\arraybackslash}m{1.5cm}}
\newcolumntype{M}{>{\centering\arraybackslash}m{1cm}}
\setlist[enumerate]{itemsep=0mm}
\DeclareMathOperator{\E}{\mathbb{E}}
\DeclarePairedDelimiter\ceil{\lceil}{\rceil}

\definecolor{ao(english)}{rgb}{0.0, 0.5, 0.0}

\title{Mimicking Evolution with Reinforcement Learning}

\author{
  João P. Abrantes\thanks{For correspondence visit http://joao-abrantes.com} \\
  Independent Researcher
  \And
  Arnaldo J. Abrantes \\
  Instituto Superior de Engenharia de Lisboa\\
  \And
  Frans A. Oliehoek\\
  Delft University of Technology
}

\begin{document}

\maketitle

\begin{abstract}
Evolution gave rise to human and animal intelligence here on Earth. We argue that the path to developing artificial human-like-intelligence will pass through mimicking the evolutionary process in a nature-like simulation. In Nature, there are two processes driving the development of the brain: evolution and learning. Evolution acts slowly, across generations, and amongst other things, it defines what agents learn by changing their internal reward function. Learning acts fast, across one’s lifetime, and it quickly updates agents’ policy to maximise pleasure and minimise pain. The reward function is slowly aligned with the fitness function by evolution, however, as agents evolve the environment and its fitness function also change, increasing the misalignment between reward and fitness. It is computationally expensive to replicate these two processes in simulation. This work proposes Evolution via Evolutionary Reward (EvER) which allows learning to single-handedly drive the search for policies with increasingly evolutionary fitness by ensuring the alignment of the reward function with the fitness function. In this search, EvER makes use of the whole state-action trajectories that agents go through their lifetime. In contrast, current evolutionary algorithms discard this information and consequently limit their potential efficiency at tackling sequential decision problems. We test our algorithm in two simple bio-inspired environments and show its superiority at generating more capable agents at surviving and reproducing their genes when compared with a state-of-the-art evolutionary algorithm.
\end{abstract}

\section{Introduction}
Evolution is the only process we know of today that has given rise to general intelligence (as demonstrated in animals, and specifically in humans). This fact has been inspiring artificial intelligence (AI) researchers to mimic biological evolution in computers for a long time. With better tools and more efficient computational methods to mimic evolution, we can increase the pace of research discoveries. In fact, having these tools would not only have a major impact in AI research but it would also speed up research across a multitude of other fields. The study of different types of replicators (the entities which are subject to evolutionary pressure, such as genes, ideas, business plans and behaviours) and their specific mechanisms for replicating makes up the core foundation for different research fields (such as biological evolution, mimetics~\cite{selfish}, economics~\cite{galor2002natural, beinhocker2006origin, nelson2009evolutionary} and sociology~\cite{dickens2000social, blueprint} correspondingly). The way a population of replicators evolves in all of these fields is based on the following universal truth: replicators that are better at surviving and self-replicating increase in numbers faster than their less capable peers. When these replicators are competing for a common limited-resource, the less capable ones eventually disappear leaving the more capable peers to dictate the world’s future. In this work we propose a reward function that allows reinforcement learning (RL) algorithms to maximise the evolutionary fitness of the agents being trained. This reward function is suitable for any \emph{open-ended} evolutionary environment, that we define as: a never-ending environment where adaptable replicators are competing for a common limited-resource to survive and replicate. In these environments, as in Nature, the fitness function (or any goal function) is not defined anywhere but simply emerges from the survival and reproduction of replicators.

In the remaining part of this introduction we 1) motivate a promising research methodology for progress in artificial general intelligence (AGI) based on convergent evolution; 2) describe how evolution changes what we learn; 3) introduce our contribution and describe how maximising a reward function can lead to the maximisation of evolutionary fitness.

\subsection{A promising methodology for progress in AGI based on Convergent Evolution}
\label{sec:method}
Convergent Evolution occurs when different species independently evolve similar solutions to solve similar problems. For example, the independent evolution of eyes has occurred at least fifty times across different species~\cite{fernald2006casting, blueprint} (most famously, the anatomy of an octopus's eye is incredibly similar to the human eye despite our common ancestor having lived more than 750 million years ago and had practically no ability to see much beyond detecting the absence or presence of light). Additionally, there is now compelling evidence that complex cognitive abilities such as love, friendship and grief have been independently evolved in social species such as elephants, dolphins, whales and humans to solve the problems that occur when individuals interact frequently with members of the same species~\cite{blueprint}. We argue that to evolve AI and have it converge to similar cognitive abilities as the ones found in nature, we need to subject artificial agents to the same problems life finds in nature. In nature, the degree of adaptability of living beings together with the level of freedom provided by their natural environment enables the growth of a wide variety of complex beings. A modified DNA can alter what an animal can sense and do by changing the animal’s body (its sensors and actuators). It can alter how an animal learns by modifying its learning system and its brain architecture. Lastly, it can also alter what knowledge the animal is born with by changing its instincts (its innate behaviours). Achieving the level of freedom of the natural environment and the level of adaptability that the DNA provides in a simulation is not feasible. We need to reduce the adaptability degree of our agents and the degree of freedom of their environment but still be able to evolve complex cognitive abilities similar to the ones found in nature. With that end, we propose the following methodology:
\begin{enumerate}
    \item Design simplified bio-inspired agent(s) to live in a simplified bio-inspired environment.
    \item Evolve these agents.
    \item Compare the artificial behaviours obtained with the natural behaviours observed in the natural environment that served as inspiration.
    \item If there is a mismatch, formulate a hypothesis that might explain it. Did the mismatch occur due to an oversimplification of the agent’s body (sensors and actuators)? Was it due to its brain architecture or its reward system? Repeat from step 1 to test the new hypothesis.
\end{enumerate}

The general idea of using first principles to build generative models, collect data from simulations and compare it to data from the real world, was proposed by Axelrod in 1997~\cite{axelrod1997advancing} and baptized as \textit{the third way of doing science} as in differing from the deductive and inductive ways of doing science.

Past work in artificial life~\cite{langton1997artificial}, evolutionary computation~\cite{eiben2003introduction} and co-evolution~\cite{reynolds1994competition} contribute towards different steps of this methodology. Our contribution is towards the 2\textsuperscript{nd} step, as our proposed reward function allows RL algorithms to maximise an agent's evolutionary fitness.

\subsection{Evolving what to learn}
\label{sec:twoloops}
\begin{figure}
  \centering
  \includegraphics[scale=0.35]{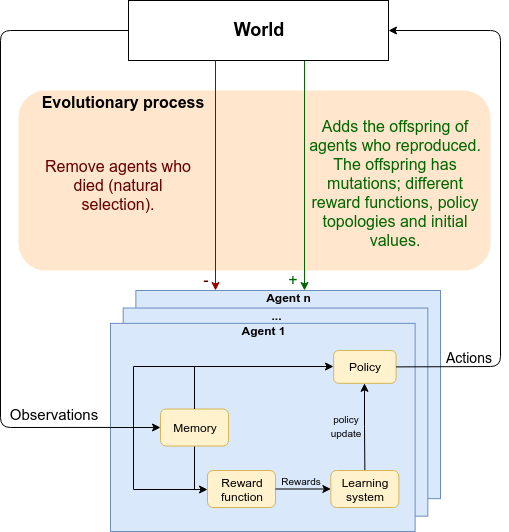}
  \caption{A diagram on how evolution and learning affect the development of the brain.}
  \label{fig:diagram}
\end{figure}

In nature, there are two different mechanisms driving the development of the brain (figure \ref{fig:diagram}). Evolution acts slowly, across generations, and amongst other things, it defines what agents learn by changing their internal reward function. Learning acts fast, across one’s lifetime. It quickly updates agents' policy to maximise pleasure and minimise pain. Combining evolution and learning has long history in AI research. The evolutionary reinforcement learning algorithm, introduced in 1991~\cite{ackley1991interactions}, makes the evolutionary process determine the initial weights of two neural networks: an action and an evaluation network. During an agent's lifetime, learning adapts the action network guided by the output of its innate and fixed (during its lifetime) evaluation network. NEAT+Q~\cite{whiteson2006evolutionary} uses an evolutionary algorithm, NEAT~\cite{stanley2002evolving}, to evolve topologies of NN and their initial weights so that they can better learn using RL. In NEAT-Q the reward function remains fixed. However, evolutionary algorithms have also been used to evolve potential-based shaping rewards and meta-parameters for RL~\cite{elfwing2008co}. 

We say that a reward function is aligned with evolution when the maximisation of the reward leads to the maximisation of the agent’s fitness. Evolution slowly aligns the reward function. However, when agents are evolved together, the behavior of some agents effect the fitness of others. As such, the fitness landscape of an individual is changing over time, as the populations evolve. Therefore, the typical internal rewards function that individuals optimize, also keep evolving over time: there is no clear fixed point to which evolution of the reward function would converge. More formally, we say that the environment from the perspective of agent $i$ is defined by the state transition distribution; $E^i \coloneqq p(s_{t+1}^i|s_t^i,a_t^i,\bm{\pi}^{-i})$. Where $\bm{\pi}^{-i}$ is the concatenation of the policies of all agents except agent $i$; $\bm{\pi}^{-i} \coloneqq \{\pi^j\}_{\forall j \neq i}$. Policy adaptation occurs by: $h^i_\pi\sim <\mathcal{L}(\mathcal{R}^i), E^i>$, where $h^i_\pi$ is the sampled history of the adaptations of policy $\pi^i$ which resulted from agent $i$ learning $\mathcal{L}(.)$ to maximise its reward function $\mathcal{R}^i$ by interacting $<.>$ with the environment. If agent $i$ is the only agent learning then $\bm{\pi}^{-i}$ is static and so is the environment $E^i$. In this case, the optimally aligned reward function is given by:
\begin{equation}
    \mathcal{R^*} = \arg\max_{\mathcal{R}^i}\E_{h^i_\pi\sim <\mathcal{L}(\mathcal{R}^i), E^i>}\mathcal{F}(h^i_\pi,E^i)
\end{equation}
Where $\mathcal{F}$ is the fitness function. The notion of an optimal reward function for a given fitness function was introduced by Singh~\cite{lewis2010rewards, singh2010intrinsically}, here we adapted his original formulation. In the general case, all agents are learning, and therefore, the environment is non-static, the fitness for $h^i_\pi$ is changing and so is the optimally aligned reward $\mathcal{R}^*$. However, in this paper, we show that in simulation it is possible to \emph{define} a fixed reward function which is always aligned, although not guaranteed to be optimally aligned, with the essence of fitness: the ability of the individual to survive and reproduce its replicators (genes).

The implementation of the described framework (figure \ref{fig:diagram}) results in a very computational expensive algorithm as it requires two loops 1) learning (the inner loop) where agents maximise their innate reward functions across their lifetimes and 2) evolution (the outer loop) where natural selection and mutation defines the reward functions for the next generation (amongst other things, such as NN topologies and initial weights). Our work allows learning to single-handedly drive the search for policies with increasingly evolutionary fitness by ensuring the alignment of the reward function with the fitness function. We can do this because our reward is extrinsic to the agent, it's given by God, and therefore, only possible within a simulation.

\subsection{Learning to maximise evolutionary fitness}
\label{sec:maximise_fitness}
The distinction between an agent and a replicator is key to understanding this paper. Formally, evolution is a change in replicator frequencies in a population (of agents) over time. The replicator is the unit of evolution, and an agent carries one or more replicators. For example, in biology, evolutionary pressure occurs at the gene level, not at the level of the individual. In this case, the replicator is the gene, and the agent is the animal, plant, bacteria, and so on. Richard Dawkins has famously described our bodies as throwaway survival machines built for replicating immortal genes~\cite{selfish}. His description illustrates well the gene-centered view of evolution~\cite{williams1966adaptation, selfish}, a view that has been able to explain multiple phenomena such as intragenomic conflict and altruism that are difficult to explain with organism-centered or group-centered viewpoints~\cite{austin2009genes, gardner2017meaning, dawkins1979twelve}. The world would be a very different place if evolution was organism-centered, it is therefore crucial, to evolve biological-like cognition, that we exert evolutionary pressure on the replicators and not on the individuals.

Evolution acts on the replicator level, but RL acts on the agent level. RL can be aligned with the evolutionary process by noting what evolution does to the agents through its selection of replicators: evolution generates agents with increasing capabilities to maximise the survival and reproduction success of the replicators they carry. 

From the replicator’s perspective, the evolutionary process is a constant competition for resources. However, from the agent’s perspective, the evolutionary process is a mix between a cooperative exercise with agents that carry some of its replicators (its family) and a competition with unrelated agents. Evolution pressures agents to engage in various degrees of collaboration depending on the degree of kinship between them and the agents they interact with (i.e. depending on the amount of overlap between the replicators they carry). This pressure for cooperation amongst relatives was named \textit{kin selection}~\cite{smith1964group} \footnote{It was informally summarised by J.B.S. Haldane that has jokingly said "I would gladly lay down my life for two brothers or eight first cousins"~\cite{dugatkin2007inclusive} (alluding to the fact that in humans, and other diploid species, siblings have in average $\frac{1}{2}$ of the genes identical by descent, whilst first cousins have in average $\frac{1}{8}$, therefore, from the genes perspective sacrificing an individual to save two siblings or eight first cousins is a fair deal, evolutionary speaking)}.

Cooperation in multi-agent reinforcement learning (MARL) is usually approached in a binary way~\cite{OpenAI_dota, vinyals2019grandmaster, kurach2019google}. Agents are grouped into teams and agents within the same team fully cooperate amongst each other whilst agents from different teams don’t cooperate at all (cooperation is either one or zero); we define this scenario as the binary cooperative setting. In this paper, we extend the concept of team to the concept of family to move from binary cooperation into continuous degrees of cooperation. We then use this family concept to propose a new RL reward function that is aligned with the fitness landscape of an open-ended evolutionary environment.

\section{Related work}
\paragraph{Arms-race with non-gradient methods} To achieve increasingly complex behaviours, agents have to face an increasingly complex task. This happens when they compete against adaptable entities. Their task gets harder every time their competitors learn something useful - an arms race is created and it drives the continued emergence of ever new innovative and sophisticated capabilities necessary to out-compete adversaries. Evolutionary Algorithms (EA) have been successfully used to co-evolve multiple competing entities (some classical examples are: Sims, 1994~\cite{sims1994evolving} and Reynolds, 1994~\cite{reynolds1994competition}). However, in sequential decision problems EA algorithms discard most of the information by not looking at the whole state-action trajectories the agents go through their lifetime. This theoretical disadvantage limits their potential efficiency to tackle sequential problems when compared with RL. Empirically, EA algorithms usually have a higher variance when compared with gradient methods~\cite{runarsson2005coevolution, lucas2006temporal, lucas2007point}.

\paragraph{Arms-race with gradient methods} With regards to gradient methods (deep learning methods in specific), impressive results have been recently achieved by training NN, through back-propagation, to compete against each other in simulated games (OpenFive~\cite{OpenAI_dota}, AlphaZero~\cite{silver2018general}, GAN~\cite{goodfellow2014generative}). More closely aligned with our proposed methodology, OpenAI has recently developed Neural MMO~\cite{suarez2019neural}, a simulated environment that captures some important properties of Life on Earth. In Neural MMO artificial agents, represented by NN, need to forage for food and water to survive in a never-ending simulation. Currently, they have presented results for agents which use RL to maximise their survival, instead of maximising the survival and reproduction success of a replicator as it happens in nature. We build on top of this work and add three important contributions: we introduce replicators (genes), give our agents the ability to reproduce, and design an RL algorithm that increases agents capability to maximise their genes survival and replication success. These are key properties of life on Earth that we must have in simulation environments if we hope to have them evolve similar solutions to the ones evolved by nature (in other words, these are key properties to achieve convergent evolution).

\paragraph{Cooperative MARL} Cooperative MARL is an active research area within RL that has been experiencing fast progress~\cite{OliehoekAmato16book, bansal2017emergent, foerster2018counterfactual}. The setting is usually described as a binary cooperation exercise where agents are grouped into teams and all team members receive the same reward. The teams may have a fixed number of members or change dynamically ~\cite{kok2005utile, oliehoek2013approximate, van2016coordinated, bohmer2019deep}. The most straightforward solution would be to train independent learners to maximise their team's reward. However, independent learners would face a non-stationary learning problem, since from their perspective the learning process of other agents is changing the dynamics of the environment. The MADDPG~\cite{lowe2017multi} algorithm tackles this problem by using a multi-agent policy gradient method with a centralised critic and decentralised actors so that training takes into account all the states and actions of the entire team but during execution each agent can act independently. More relevant to our work, factored value functions\cite{guestrin2002multiagent} such as Value Decomposition Networks (VDN)~\cite{sunehag2017value}, Q-Mix~\cite{rashid2018qmix} and QTRAN~\cite{son2019qtran} use different methods to decompose the team's central action-value function into the decentralised action-value functions of each team member. We build on top of VDN (which is further explained in section \ref{sec:vdn}) to extend the concept of team to the concept of family and introduce continuous degrees of cooperation.

\section{Background}
\label{sec:background}
\paragraph{Reinforcement Learning} We recall the single agent fully-observable RL setting~\cite{sutton1998introduction}, where the environment is typically formulated as a Markov decision process (MDP). At every time step, $t=1, 2, \dots$, the agent observes the environment's state $s_t \in \mathcal{S}$, and uses it to select an action $a_t \in \mathcal{A}$. As a consequence, the agent receives a reward $r_t\in\mathcal{R}\subset \mathbb{R}$ and the environment transitions to the state $s_{t+1}$. The tuple ($s_{t+1}, r_t$) is sampled from the static probability distribution $p: \mathcal{S} \times \mathcal{A} \to \mathcal{P}(\mathcal{S}\times \mathcal{R})$ whilst the actions $a_t$ are sampled from the parametric policy function $\pi_\theta: \mathcal{S} \to \mathcal{P}(\mathcal{A})$:
\begin{equation}
    s_{t+1}, r_t \sim p(s_{t+1}, r_{t}|s_t, a_t), \quad a_t \sim \pi_\theta(a_t|s_t)
\end{equation}
The goal of the agent is to find the optimal policy parameters $\theta^*$ that maximise the expected return  $\Bar{R}=\mathbb{E}[\sum_{t=0}^\infty\gamma^t r_t]$, where $\gamma$ is the discount factor. In the more general framework, the state is only partially observable, meaning that the agent can not directly observe the state but instead it observes $o_t \in \mathcal{O}$ which is typically given by a function of the state. In this situation, the environment is modelled by a partial observable Markov decision process (POMDP) and the policy usually incorporates past history $h_t=a_0o_0r_0, \dots, a_{t-1}o_{t-1}r_{t-1}$.

\paragraph{Q-Learning and Deep Q-Networks} The action-value function $Q^\pi$ gives the estimated return when the agent has the state history $h_t$, executes action $a_t$ and follows the policy $\pi$ on the future time steps. It can be recursively defined by $Q^{\pi}(h_t, a_t) = \mathbb{E}_{s_{t+1},r_{t} \sim p}\big[r_{t}+\gamma \mathbb{E}_{a_{t+1} \sim \pi}[Q^\pi(h_{t+1}, a_{t+1})]\big]$. Q-learning and Deep Q-Networks (DQN)~\cite{mnih2015human} are popular methods for obtaining the optimal action value function $Q^*$. Once we have $Q^*$, the optimal policy is also available as $\pi^*=\arg\max_{a_t} Q^*(h_t,a_t)$. In DQN, the action-value function is approximated by a deep NN with parameters $\theta$. $Q_\theta^*$ is found by minimising the loss function:
\begin{equation}
    \mathcal{L}_t(\theta) = \mathbb{E}_{h_t,a_t,r_t,h_{t+1}}[(y_t-Q_\theta^\pi(h_t, a_t))^2],\quad \text{where } y_t=r_t+\gamma \max_{a'}Q_\theta^\pi(a_{t+1}, h_{t+1})
\end{equation}
Where $\pi$ is the $\epsilon$-greedy policy which takes action $\arg\max_{a_t} Q^\pi(a_t, h_t)$ with probability $1-\epsilon$, and takes a random action with probability $\epsilon$.

\paragraph{Multi-Agent Reinforcement Learning} In this work, we consider the MARL setting where the underlying environment is modelled by a partially observable Markov game~\cite{hansen2004dynamic}. In this setting, the environment is populated by multiple agents which have individual observations and rewards and act according to individual policies. Their goal is to maximise their own expected return.

\paragraph{Binary Cooperative MARL and VDN}
\label{sec:vdn}
Our work builds on VDN~\cite{sunehag2017value}, which was designed to address the binary cooperative MARL setting. In this setting, the agents are grouped into teams and all the agents within a team receive the same reward. VDN's main assumption is that the joint action-value function of the whole team of cooperative agents can be additively decomposed into the action-value functions across the members of the team.
\begin{equation}
    \label{eq:vdn_team}
    Q^\mathcal{T}((h_t^1, h_t^2, \dots, h_t^{|\mathcal{T}|}), (a_t^1, a_t^2, \dots, a_t^{|\mathcal{T}|})) \approx \sum_{i \in \mathcal{T}}\tilde{Q}^i(h_t^i,a_t^i)
\end{equation}
Where $\mathcal{T}$ is the set of agents belonging to the team,  and $\tilde{Q}^i(h_t^i,a_t^i)$ is the value function of agent $i$ which depends solely on its partial observation of the environment and its action at time $t$. $\tilde{Q}^i$ are trained by back-propagating gradients from the Q-learning rule through the summation.
\begin{equation}
    g_i = \nabla \theta_i(y^\mathcal{T}_t-\sum_{i \in \mathcal{T}}\tilde{Q}(h^i_t,a^i_t|\theta_i))^2, \qquad
    y^\mathcal{T}_t = r^\mathcal{T}_t+\gamma \sum_{i \in \mathcal{T}}\max_{a^i_{t+1}}\tilde{Q}(h^i_{t+1},a^i_{t+1}|\theta_i)
\end{equation}
Where $\theta_i$ are the parameters of $\tilde{Q}^i$, $g_i$ is its gradient and $r^\mathcal{T}_t$ is the reward for the team $\mathcal{T}$ at the time instant $t$. Note that even though the training process is centralised, the learned agents can be deployed independently, since each agent acting greedily with respect to its own $\tilde{Q}^i$ will also maximise its team value function $\arg\max_{a^i_t} Q^{\mathcal{T}}_t(\dots) \approx \arg\max_{a^i_t} \tilde{Q}^i(h^i_t, a^i_t)$.

\section{Evolution via Evolutionary Reward}
In this section, we propose a reward function that enables RL algorithms to search for policies with increasingly evolutionary success. We call this reward the evolutionary reward because it is always aligned with the fitness function (see section \ref{sec:twoloops} for a definition of this alignment), and therefore, we don't need to go through the expensive process of aligning the agents' reward functions through evolution. We also propose a specific RL algorithm that is particularly suited to maximise the evolutionary reward in open-ended evolutionary environments (however other RL algorithms could also be used).

\paragraph{Evolutionary reward}
\label{sec:reward}
The evolutionary reward of an agent is proportional to the number of copies its replicators have in the world's population. Maximising this reward leads to the maximisation of the survival and reproduction success of the replicators an agent carries. We start by defining the kinship function between a pair of agents $i$ and $j$, who carry $N$ replicators represented by the integer vectors $\bm{g}^i$ and $\bm{g}^j$ (we chose to use $\bm{g}$ for genome, which in biology is the set of genes (replicators) an agent carries):
\begin{equation}
    \label{def:kinship}
    %\begin{split}
    %k\colon \mathbb{Z}^N\times \mathbb{Z}^N \to [0, 1]&,\quad (\bm{g}^i, \bm{g}^j) \mapsto k(\bm{g}^i, \bm{g}^j)\\
    k\colon \mathbb{Z}^N\times \mathbb{Z}^N \to [0, 1], \qquad
    k(\bm{g}^i, \bm{g}^j) = \frac{1}{N}\sum_{p=1}^N \delta_{g^i_p, g^j_p}
    %\end{split}
\end{equation}
Where $\delta_{g^i_p, g^j_p}$ is the Kronecker delta which is one if $g^i_p=g^j_p$ and zero otherwise%\footnote{The kinship function defined here, is different from the kinship coefficient defined in biology. In biology, the kinship coefficient is defined as the probability of an allele selected randomly from individual $i$ and an allele selected from the same "position" ($p$) from an individual $j$, being identical and from the same ancestor~\cite{lange2003mathematical}.On the other hand, the kinship function defined here does not care if the replicators came from the same ancestor or not as long as they are identical.}%
. When agent $i$ is alive at time $t+1$, it receives the reward:
\begin{equation}
    \label{eq:reward}
    r_t^i = \sum_{j \in \mathcal{A}_{t+1}}{k(\bm{g}^i, \bm{g}^j)}
\end{equation}
Where $\mathcal{A}_{t+1}$ is the set of agents alive at the instant $t+1$. Note that since agent $i$ is alive at $t+1$, $\mathcal{A}_{t+1}$ includes agent $i$. $T^i-1$ is the last time step that agent $i$ is alive and so, at this instant, the agent receives its final reward which is proportional to the discounted sum of the number of times its genes will be present on other agents after its death:
\begin{equation}
    \label{eq:final_reward}
    r_{T^i-1}^i = \sum_{t=T^i}^\infty \gamma^{t-T^i}\sum_{j \in \mathcal{A}_t}{k(\bm{g}^i, \bm{g}^j)}
\end{equation}

With this reward function, the agents are incentivised to maximise the survival and replication success of the replicators they carry. In the agent-centered view, the agents are incentivised to survive and replicate, but also to help their family (kin) survive and replicate; and to make sure that when they die their family is in a good position to carry on surviving and replicating.

The discount factor, $\gamma$, needs to be in the interval $[0, 1[$ to ensure the final reward remains bounded. Due to the exponential discounting we can compute the final reward up to an error of $\epsilon$ by summing over a finite period of time denoted by the effective horizon ($h_e$). To see how to compute the $h_e$ for a given environment and $\epsilon$ see the appendix \ref{sec:he}. We can now use Q-learning to train agents with this evolutionary reward. However, in the next section we introduce a more practical algorithm that allows us to estimate the final reward without having to simulate the environment forward for $h_e$ iterations.

\paragraph{Evolutionary Value-Decomposition Networks} We propose Evolutionary Value-Decomposition Networks (E-VDN) as an extension of VDN (explained in detail in section \ref{sec:vdn}) from the binary cooperative setting with static teams to the continuous cooperative setting with dynamic families. E-VDN helps us reduce the variance of the value estimation and allows us to estimate the final evolutionary reward without having to simulate the environment forward for $h_e$ iterations.

Within a team, each agent fully cooperates with all the other members of the team, and it does not cooperate at all with any agent outside of the team. Moreover, if $a$ and $b$ are members of the same team and $c$ is a member of $a$'s team then $c$ and $b$ are also in the same team. Within a family, the degrees of cooperation amongst its members depends on their kinship degree (which can be any real number from 0 to 1). Also, if $a$ and $b$ are members of the same family and $c$ is part of $a$'s family, $c$ is not necessarily part of $b$'s family.

Each agent $i$ sees the members of its family from an unique perspective, based on the kinship degree it shares with them. In E-VDN, each agent $i$ has a joint action-value function, $Q^i$. E-VDN assumes $Q^i$ can be composed by averaging the action-value functions across the members of $i$'s family weighted by their kinship with agent $i$ (this extends VDN's assumption, given by \eqref{eq:vdn_team}, to the continuous cooperative setting):
\begin{equation}
    \label{eq:maeq}
    Q^i((h_t^1, h_t^2, \dots, h_t^{|\mathcal{A}_t|}), (a_t^1, a_t^2, \dots, a_t^{|\mathcal{A}_t|})) \approx \frac{1}{n^i_t}\sum_{j \in \mathcal{A}_t}k(\bm{g}^i,\bm{g}^j)\tilde{Q}^j(h_t^j,a_t^j)
\end{equation}
Where $n^i_t$ is a normalisation coefficient defined as $n^i_t=\sum_{j \in \mathcal{A}_t}k(\bm{g}^i,\bm{g}^j)$. Composing $Q^i$ with an average, instead of a sum, is necessary as E-VDN allows the number of value functions contributing to the composition to vary as the family gets bigger or smaller (agents born and die). This averaging allows us to incorporate the local observations of each family member and reduce variance in the value estimation. 

More importantly, E-VDN allows us to deal with the difficulty of estimating the terminal reward \eqref{eq:final_reward} in a particularly convenient way. As is clear from its definition \eqref{eq:final_reward}, the terminal reward is the expected sum (over time) of kinship that agent $i$ has with other agents $j$ after its death. The key idea is to note that this value ($r^i_{T^i-1}$) can be approximated by the Q-value of other agents $j$ that are close to (have high kinship with) agent $i$:
\begin{equation}
    \hat{r}^i_{T^i-1} = \left\{
	    \begin{array}{ll}
		    \frac{1}{n^i_{T^i}}\sum_{j\in \mathcal{A}_{T^i}}k(\bm{g}^i,\bm{g}^j)\tilde{Q}^j_{T^i}(\dots)\approx Q^i_{T^i}(\dots) & \mbox{if } n^i_{T^i} > 0 \\
		    0 & \mbox{if } n^i_{T^i} = 0
	    \end{array}
    \right. \label{eq:final_rew_estimate}
\end{equation}
The final reward is zero if, and only if, at the time of its death the agent has no surviving family. 

Each $\tilde{Q}^i$ is trained by back-propagating gradients from the Q-learning rule:
\begin{equation}
    \label{eq:gradient}
    	g^i_t = \nabla \bm{\theta_i}(y^i_t-\frac{1}{n^i_t}\sum_{j \in \mathcal{A}_t}k(\bm{g}^i,\bm{g}^j)\tilde{Q}^j(h^j_t, a^j_t|\tilde{\theta}_j))^2 \approx \nabla \bm{\theta_i}(y^i_t-Q^i_t(\dots|\bm{\theta_i}))^2
\end{equation}
Where $\bm{\theta}_i$ is the concatenation of all the parameters $\tilde{\theta}_j$, used in each $\tilde{Q}^j$, contributing to the estimation of $Q^i$; i.e. $\bm{\theta}_i \coloneqq \{ \tilde{\theta}_j \}_{j~\text{s.t.}~k(\bm{g}^i, \bm{g}^j) > 0}$. Note that $\tilde{Q}^i$ are neural networks with parameters $\tilde{\theta}_i$ and $Q^i$ is simply the average stated in \eqref{eq:maeq}.

The learning targets $y^i_t$ are given by:
\begin{align}
    y^i_t &= \left\{
	    \begin{array}{ll}
		    r^i_t+\gamma \max_{\bm{a}_{t+1}}Q^i_{t+1}(\dots)) & \mbox{if } t < T^i-1 \\
		    \hat{r}^i_{T^i-1} & \mbox{if } t = T^i-1
	    \end{array}
    \right. \label{eq:learning_targets}
\end{align}
$r^i_t$ is the evolutionary reward \eqref{eq:reward} and $\hat{r}^i_{T^i-1}$ is the estimate of the final evolutionary reward  \eqref{eq:final_rew_estimate}. We don't use a replay buffer in our training (which is commonly used in DQN) due to the non-stationary of multi-agent environments (more about this in the appendix \ref{sec:replay_buffer}).

Since the joint action-value $Q^i$ increases monotonically with increasing $\tilde{Q}^i$, an agent acting greedily with respect to its action-value function will also act greedily in respect to its family action-value function: $\arg\max_{a^i_t}Q^i_t(\dots) \approx \arg\max_{a^i_t}\tilde{Q}^i(h^i_t, a^i_t)$. 

\section{Experimental Setup}
We want to test two hypotheses: 1) E-VDN is particularly well suited to make agents climb the fitness landscape in open-ended evolutionary environments; 2) E-VDN is able to increase the evolutionary fitness of agents in non-binary cooperative environments. To test these hypotheses, we introduce a binary and non-binary cooperative setting: the asexual (binary) and the sexual (non-binary) environments. In this section, we give a quick overview of these two multi-agent environments, as well as details of the network architectures and the training regime. For a more complete description of the environments, you can refer to the appendix \ref{sec:env}.

\begin{figure}
  \centering
  \includegraphics[scale=0.15]{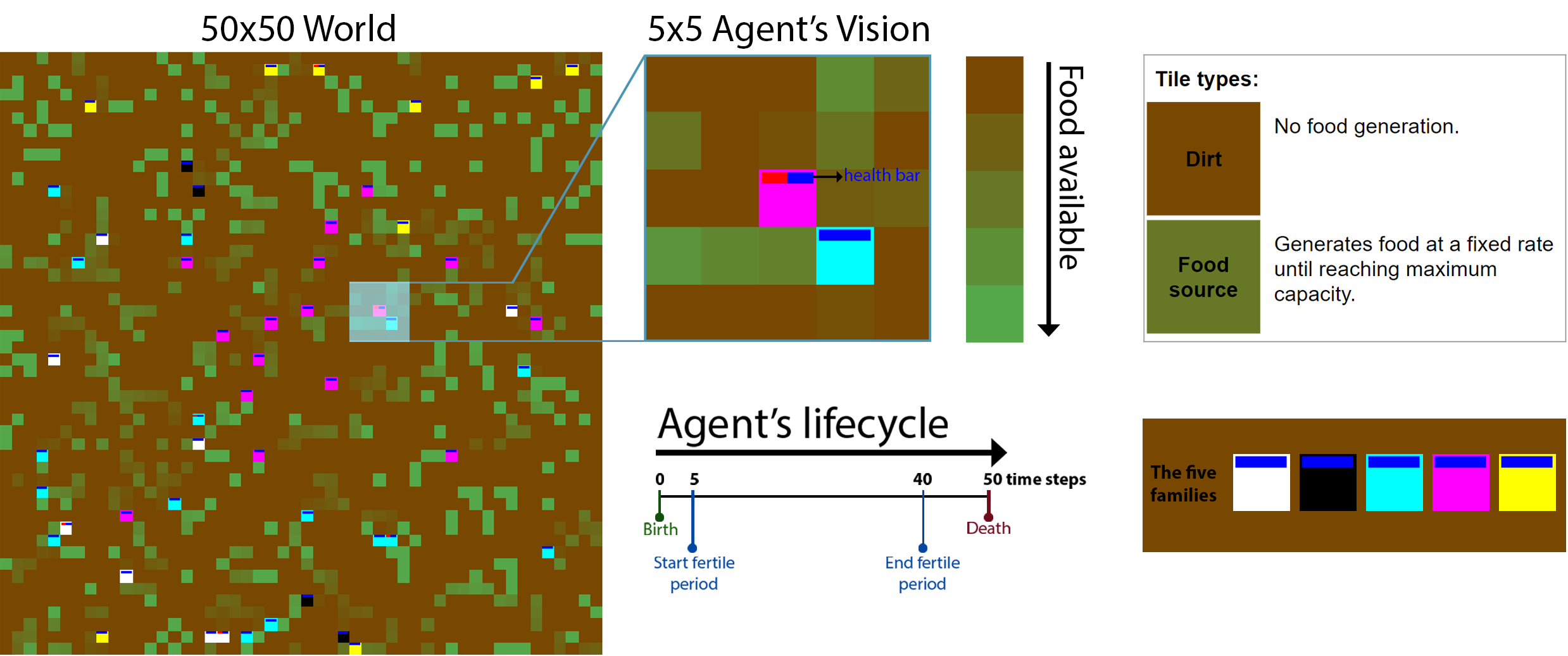}
  \caption{The asexual environment.}
  \label{fig:env}
\end{figure}

\paragraph{The Asexual Environment} The asexual environment is a 2-dimensional grid world, which is initialised with five agents carrying five unique genomes (figure \ref{fig:env}). At each time step, each agent may move one step and produce an attack to another agent in an adjacent tile. When an agent moves to a tile with food it collects all the food available in it. If an agent chooses to produce an attack, it decreases its victim's health by one point, if the victim's health reaches zero it dies and 50\% of its collected food is captured by the attacker. The food is used to survive (one unit of food must be consumed every time step to remain alive), and to reproduce. When agents are within their fertile age and they have stored enough food, they reproduce themselves asexually and give birth to an agent carrying an exact copy of their genome. Each genome has only a single gene and there are no mutations. These rules make the cooperation between agents binary, agents either fully-cooperate (they have the exact same genome) or they don't cooperate at all (their genome has no overlap). Note that despite this environment being in the binary cooperative setting, the VDN algorithm can not be directly applied to it since the "team" sizes increase and decrease as agents born and die, moreover, VDN would have to estimate the agent's final reward by simulating the environment forward for the effective horizon. E-VDN solves these problems by decomposing the team action-value function with an average instead of a sum, and estimating the final reward of a dying agent by using the value-functions of the remaining family members.
\paragraph{The Sexual Environment} The sexual environment has the same rules as the asexual environment with the difference that the agents now have 32 genes in their genome and they reproduce sexually. When two fertile agents are adjacent, they give birth to an agent who's genome is composed by two halves of the genes of each parent, selected randomly. There are no genders, any agent can reproduce with any other agent. These rules give rise to different levels of collaboration (33 levels of cooperation to be exact, from 0 to 1 in steps of $\frac{1}{32}$). 

\paragraph{Policy} Each agent observes a 5x5 square crop of the surrounding state (figure \ref{fig:env}). The agent sees six features for every visible tile; i.e.\ the input is a 5x5x6 tensor. This includes two features corresponding to tile properties (food available and whether it is occupied or not) and four features corresponding to the occupying agents' properties (age, food stored, kinship and health). Besides these local inputs, each agent also observes its absolute position, family size and the total number of agents in the world. We intend to remove these extra inputs in future work as we provide agents with memory (we're currently providing our policy with $o^i_t$ instead of $h^i_t$). The NN has ten outputs (five movement actions with no attack and five movement actions with an attack). In this work, we used two different feed forward architectures: one is simply a fully connected NN with three hidden layers and 244 288 parameters in total, the other architecture is composed by convolutional and dense layers and it is much smaller containing only 23 616 parameters. The smaller NN was used to compare our algorithm with an evolutionary algorithm which doesn't scale well to larger networks.

\paragraph{Training details} In the asexual environment, we train five different policies (with the same architecture but different weights) simultaneously. At each training episode, we sample five policies with replacement and assign each one to one of the five unique genomes. We do this, to force each policy to interact with all other policies (including itself), increasing their robustness in survival and reproduction. During the test episodes, no sampling occurs, each policy is simply assigned to each unique genome. The training episodes had a length between 450 and 550 (note that the reward is computed as if there was no episode end), and the test episodes had a length of 500 steps.

In the sexual environment, due to the large number of unique genomes, it is unfeasible to assign a unique policy to each unique genome. To keep things simple, we chose to use only one policy in this environment. In this work, the genome does not directly encode the policy, however, we think it would be interesting to do that in future work. For example in tabular RL the policy could be a table of integers (the actions to take at every discrete state) and the genome could be that same table of integers, EvER would then evolve the agents' genomes and policies.

\paragraph{Evolution Strategies} In the asexual environment, we compare the E-VDN algorithm with a popular ES algorithm. ES algorithms optimise an agent’s policy by sampling policy weights from a multivariate Gaussian distribution, evaluating those weights on the environment, giving them a fitness score and updating the Gaussian distribution parameters so that the next samples are more likely to achieve a higher fitness. There are a few different methods on how to update the distribution parameters, we chose to use CMA-ES~\cite{hansen2003reducing} because it has been successful in optimising NN for a wide range of sequential decision problems~\cite{heidrich2008evolution, heidrich2008variable, heidrich2009hoeffding}. However, note that CMA-ES was not designed for multi-agent settings where the fitness function landscape changes as the other agents learn. Nevertheless, we used five independent multivariate Gaussians distributions each one associated with a unique gene and each one being updated by the CMA-ES algorithm. In the beginning, when the agents can not survive for long, the fitness function is given by the total sum of family members along time, once the agents learn how to survive and reproduce we change the fitness function to be the number of family members at the end of an episode with 500 steps. Since the CMA-ES algorithm computation time grows quadratic with the number of parameters, O($N^2$), we had to use the smaller NN for this comparison. This algorithm was not applied to the sexual environment due to the large number of unique genomes available in this environment. The algorithm was implemented using an available \textit{python} library~\cite{hansen2019pycma}.

\paragraph{Evaluation Metrics} In our simple environments, fitter policies can use the environment resources more efficiently and increase their population size to larger numbers. Therefore, to evaluate the performance of the algorithms in generating increasingly fitter species we track the average population size along training time. 

\section{Results}
\begin{figure}
  \centering
  \includegraphics[scale=0.37]{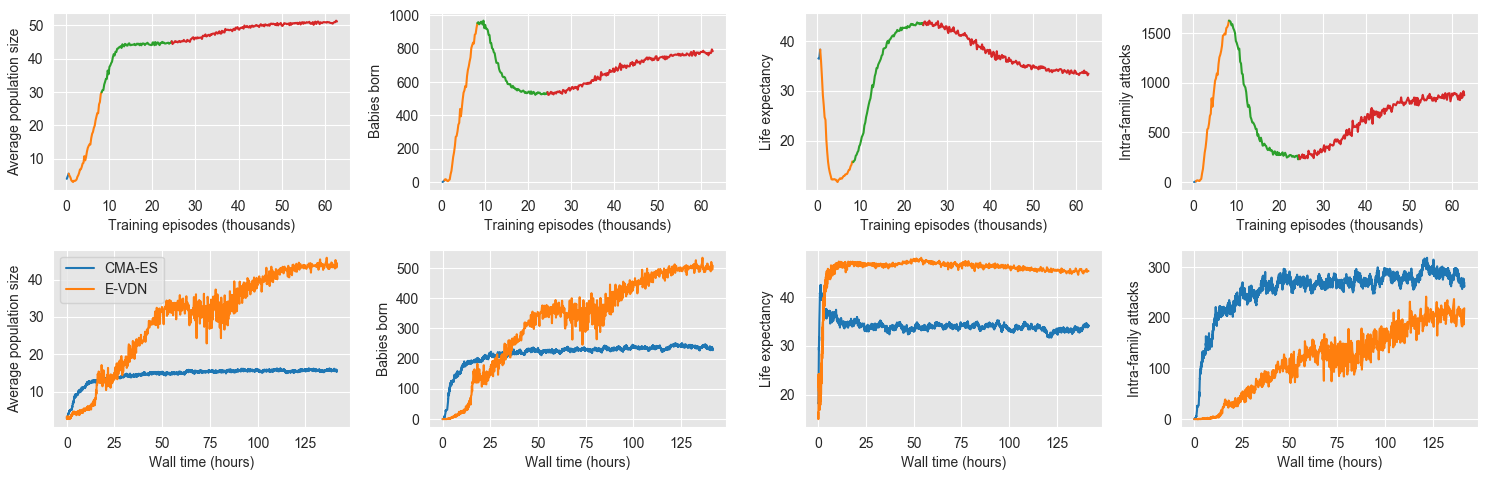}
  \caption{(1\textsuperscript{st} row) Results obtained using E-VDN with the larger NN, each point was obtained by averaging 20 test episodes. The different colours correspond to different eras. This plot was generated with a denser version of the evolutionary reward (more details on the appendix \ref{sec:denser_reward}). (2\textsuperscript{nd} row) Results obtained using CMA-ES and E-VDN algorithms with the smaller NN and the standard evolutionary reward \eqref{eq:reward}. Both algorithms were trained with 20 CPUs each.
}
  \label{fig:history}
\end{figure}

Training agents with E-VDN generates quite an interesting evolutionary history. Throughout the asexual environment history, we found four distinct eras where agents engage in significantly distinct behaviour patterns (1\textsuperscript{st} row of fig. \ref{fig:history}). In the first era (the blue line - barely visible in the figure), the agents learned how to survive, and through their encounters with the other founding agents, they have learnt that it was always (evolutionary) advantageous to attack other agents. In the second era (orange line), the agents' food-gathering skills increased to a point where they started to reproduce. In this era, the birth-rate and population numbers increased fast. However, with the extra births, intra-family encounters became more frequent, and intra-family violence rose to its all-time maximum driving the average life span down. This intra-family violence quickly decreased in the third era (green line), as agents started to recognize their kin. Kin detection allowed for selective kindness and selective violence, which took the average life span to its all-time maximum. Finally, in the fourth era (red line), agents learned how to sacrifice their lives for the future of their family. Old infertile agents started allowing the younger generation to eat them without retaliation. Through this cannibalism, the families had found a system for wealth inheritance. A smart allocation of the family's food resources in the fitter generation led to an increase in the population size with the cost of a shorter life span. This behaviour emerges because the final reward \eqref{eq:final_reward} incentivises agents to plan for the success of their genes even after their death. This behaviour is further investigated in the appendix \ref{sec:app_can}. These results show that optimising open-ended evolutionary environments with E-VDN does indeed generate increasingly complex behaviours.

\begin{figure}
  \centering
  \includegraphics[scale=0.37]{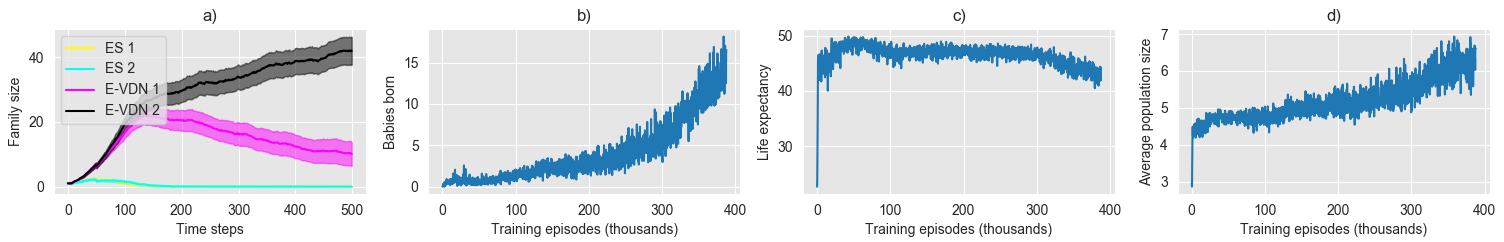}
  \caption{a) (CMA-ES vs E-VDN) Average and 95\% confidence interval of two CMA-ES and two E-VDN family sizes computed over 90  episodes. b, c and d) the macro-statistics obtained in the sexual environment. To speed up the training we used the smaller NN and the denser evolutionary reward described in the appendix \ref{sec:denser_reward}.}
  \label{fig:cannibalism}
\end{figure}

The 2\textsuperscript{nd} row of figure \ref{fig:history}, shows the macro-statistics obtained by training the smaller NN with CMA-ES and E-VDN. From the figure, we observe that E-VDN is able to produce a larger population of agents with a longer life-span and a higher birth rate. A small population means that many resources are left unused by the current population, this creates an opportunity for a new and more efficient species to collect the unused resources and multiply its numbers. These opportunities are present in the CMA-ES environment, however the algorithm could not find them, which suggests that E-VDN is better at finding the way up the fitness landscape than CMA-ES. \href{https://youtu.be/FSpQ2wNgCW8}{Video 1}, shows that each family trained with CMA-ES creates a swarm formation in a line that moves around the world diagonally. When there is only one surviving family, this simple strategy allows agents to only step into tiles that have reached their maximum food capacity. However, this is far from an evolutionarily stable strategy~\cite{smith1973logic} (ESS; i.e.\ a strategy that is not easily driven to extinction by a competing strategy), as we verify when we place the best two families trained with CMA-ES on the same environment as the best two E-VDN families and observe the CMA-ES families being consistently driven quickly to extinction by their competition (fig. \ref{fig:cannibalism}.a).

In the sexual environment, figures \ref{fig:cannibalism}.b,c and d show that E-VDN has been able to train a policy that consistently improves the survival success of the founding genes. Note, that this environment is much harder than the previous one. To replicate, agents need to be adjacent to other agents. In the beginning, all agents are unrelated making it dangerous to get adjacent to another agent as it often leads into attacks, but it is also dangerous to get too far away from them since with a limited vision it is hard to find a fertile mate once they lose sight of each other. \href{https://youtu.be/xyttCW93xiU}{Video 2} shows a simulation of the evolved policy being run on the sexual environment, it seems that agents found a way to find mates by moving to a certain region of the map (the breeding ground) once they are fertile.

\section{Conclusion \& Future Work}
This paper has introduced an evolutionary reward function that when maximised also maximises the evolutionary fitness of the agent. This allows RL to be used as a tool for research of open-ended evolutionary systems. To implement this reward function, we extended the concept of team to the concept of family and introduce continuous degrees of cooperation. Future work will be split into three independent contributions: 1) Encode the agents' policy directly in their genome (e.g. for discrete state-actions spaces a table of integers can both represent the policy and the genome); 2) Explore a different reward function that makes agents maximise the expected geometric growth rate of their replicators. We call this the Kelly reward $r^k_t=\log\frac{r_t}{r_{t-1}}$ ($r_t$ is defined in eq. \ref{eq:reward}), inspired by the Kelly Criterion~\cite{kelly2011new} (KC), an investment strategy that guarantees higher wealth, after a series of bets, when compared with any other betting strategy in the long run (i.e.\ when the number of bets tends to infinity). In fact, the field of economics (which includes investments) is often seen from the evolutionary perspective~\cite{beinhocker2006origin}. More wealth allows investors to apply their investment policy at a larger scale, multiplying its replication ability. The same can be said about genes, more agents carrying those genes allows the genes' policy to be applied at a larger scale, multiplying its replication ability. Moreover, investors can enter an absorption state from where there is no escape (bankruptcy), and the same happens with genes (extinction). The KC results from maximising the expected geometric growth rate of wealth, as opposed to the usual goal of maximising the expected wealth which, in some cases, may lead to an investment strategy that tends to give bankruptcy with probability of one as the number of bets grows to infinite (see the St. Petersburg paradox for an example); 3) Following our proposed methodology for progress in AI (section \ref{sec:method}), we will research the minimum set of requirements to emerge natural cognitive abilities in artificial agents such as identity awareness and recognition, friendship and hierarchical status. 
\bibliographystyle{plainnat}
\bibliography{ever.bib}

\begin{appendices}
\section{Videos}
- \href{https://youtu.be/FSpQ2wNgCW8}{Video 1} (\url{https://youtu.be/FSpQ2wNgCW8})\\
- \href{https://youtu.be/xyttCW93xiU}{Video 2} (\url{https://youtu.be/xyttCW93xiU})

\section{Environment}
\label{sec:env}
In this section, we go through the game loop of the environments summarised in the main article. Both the asexual and sexual environment have the same game loop, their only difference is in the way agents reproduce and in the length of the genome agents carry (the genome has a single gene in the asexual environment and 32 genes in the sexual one). The states of the tiles and agents are described in table \ref{tab:states}.

\begin{table}[h]
    \centering
    \begin{tabular}{ lc|lc } 
     \toprule
     %\rowcolor{lightgray}
     \multicolumn{2}{c}{Tile state} & \multicolumn{2}{c}{Agent state} \\
     \midrule
     Type & Boolean (food source/dirt) & Position (x,y) & Integer, Integer \\ 
     Occupied & Boolean & Health & Integer  \\ 
     Food available & Float & Age & Integer \\
     && Food stored & Float \\
     && Genome & Integer Vector\\
     \bottomrule
    \end{tabular}
    \caption{The state of the tiles and agents.}
    \label{tab:states}
\end{table}

We now introduce the various components of the game loop:
\paragraph{Initialisation} The simulation starts with five agents, each one with a unique genome. All agents start with age 0 and $e$ units of food (the endowment). The environments are never-ending. Table \ref{tab:init} describes the configuration used in the paper.

\begin{table}[h]
    \centering
    \begin{tabular}{ |L|M|L|L|c|M|L|L| } 
     \hline
     Endowment ($e$) & Initial health & Start of fertility age & End of fertility age & Longevity & World size & Food growth rate ($f_r$) & Maximum food capacity ($c_f$) \\
     \hline
     10 & 2 & 5 & 40 & 50 & 50x50 & 0.15 & 3\\
     \hline
    \end{tabular}
    \caption{Configuration of the environment used in the paper.}
    \label{tab:init}
\end{table}

\paragraph{Food production} Each tile on the grid world can either be a food source or dirt. Food sources generate $f_r$ units of food per iteration until reaching their maximum food capacity ($c_f$).

\paragraph{Foraging} At each iteration, an agent can move one step to North, East, South, West or choose to remain still. When an agent moves to a tile with food it collects all the available food in it. The map boundaries are connected (e.g.\ an agent that moves over the top goes to the bottom part of the map). Invalid actions, like moving to an already occupied tile, are ignored. 

\paragraph{Attacking} At each iteration, an agent can also decide to attack a random adjacent agent: this is an agent within one step to N, E, S or W. Each attack takes 1 unit of health from the victim’s. If the victim’s health reaches zero, it dies, and the attacker will “eat it” and receive 50\% of its food reserves. 

\paragraph{Asexual Reproduction} An agent is considered fertile if it has accumulated more than twice the amount of food it received at birth (i.e. twice its endowment $e$) and its age is within a given fertile age. The fertile agent will give birth once they have an empty tile nearby, when that happens the parent transfers $e$ units of food to its newborn child. The newborn child will have the same genome has its parent.

\paragraph{Sexual Reproduction} An agent is considered fertile if it has accumulated more than the amount of food it received at birth and its age is within a given fertile age. The fertile agent will give birth once it is adjacent to another fertile agent and one of them has an empty tile nearby, when that happens each parent transfers $\frac{e}{2}$ units of food to its newborn child. A random half of the newborn's genes come from the first parent, and the second half comes from the second parent.

\paragraph{Game loop} At every iteration, we randomise the order at which the agents execute their actions. Only after all the agents are in their new positions, the attacks are executed (with the same order as the movement actions). The complete game loop is summarized in the next paragraph.

At each iteration, each agent does the following:
\begin{itemize}
    \item Execute a movement action: Stay still or move one step North, East, South or West.
    \item Harvest: Collect all the food contained in its tile.
    \item Reproduce: Give birth to a child using asexual or sexual reproduction (see their respective sections).
    \item Eat: Consume a unit of food.
    \item Age: Get one year older.
    \item Die: If an agent's food reserves become empty or it becomes older than its longevity threshold, then it dies. 
    \item Execute an attack action: After every agent has moved, harvested, reproduced, eaten and aged the attacks are executed. Agents that reach zero health get eaten at this stage.
\end{itemize}
Additionally, at each iteration, each food source generates $f_r$ units of food until reaching the given maximum capacity ($c_f$). 

\section{Algorithm details}
\subsection{Effective time horizon} 
\label{sec:he}
We want to find the number of iterations ($h_e$) that guarantee an error between the estimate of the final reward and the actual final reward to be less or equal than a given $\epsilon$, $|r^i_{T^i-1}-\hat{r}^i_{T^i-1}| \leq \epsilon$.\\
Remember that the final reward is given by:
$$r^i_{T^i-1} = \sum_{t=T^i}^\infty \gamma^{t-T^i}\sum_{j \in \mathcal{A}_t}{k(\bm{g}^i, \bm{g}^j)}=\sum_{t'=0}^\infty \gamma^{t'}k^i_{t'}$$

Where $t'=t-T^i$ and $k^i_{t'} = \sum_{j\in \mathcal{A}_{t'}}k(\bm{g}^i, \bm{g}^j)$. The estimate of the final reward is computed with the following finite sum $\hat{r}^i_t=\sum_{t'=0}^{h_e-1} \gamma^{t'}k^i_{t'}$.\\

Note that $k^i_t$ is always positive so the error $r^i_{T^i-1}-\hat{r}^i_{T^i-1}$ is always positive as well. To find the $h_e$ that guarantees an error smaller or equal to epsilon we define $r_b$ as the upper bound of $k^i_t$ and ensure that the worst possible error is smaller or equal to epsilon:
\begin{align}
    \sum_{t'=0}^{\infty} \gamma^{t'}r_b-\sum_{t'=0}^{h_e-1} \gamma^{t'}r_b &\leq \epsilon\\
    \frac{r_b}{1-\gamma}-r_b\frac{1-\gamma^{h_e}}{1-\gamma} &\leq \epsilon\\
    \frac{r_b\gamma^{h_e}}{1-\gamma} &\leq \epsilon\\
    h_e\log{\gamma} &\leq \log{\frac{\epsilon(1-\gamma)}{r_b}}\\
    h_e &\leq \frac{\log{\frac{\epsilon(1-\gamma)}{r_b}}}{\log{\gamma}}
\end{align}

We go from (1) to (2) by using the known convergences of geometric series: $\sum_{k=0}^\infty ar^k = \frac{a}{1-r}$ and $\sum_{k=0}^{n-1} ar^k = a\frac{1-r^n}{1-r}$ for $r < 1$. Since $h_e$ needs to be a positive integer we take the ceil $h_e = \ceil*{\frac{\log{\frac{\epsilon(1-\gamma)}{r_b}}}{\log{\gamma}}}$ and note that this equation is only valid when $\frac{\epsilon(1-\gamma)}{r_b}<1$. For example, an environment that has the capacity to feed at most 100 agents has an $r_b=100$ (which is the best possible reward when the kinship between every agent is 1). If we use $\epsilon=0.1$ and $\gamma=0.9$ then $h_e=88$.

\subsection{Experience buffer}
\label{sec:replay_buffer}
When using Q-learning methods with DQN, as we are, it’s common practice to use a replay buffer. The replay buffer stores the experiences ($s_t, a_t, r_t, s_{t+1}$) for multiple time steps $t$. When training, the algorithm randomly samples experiences from the replay buffer. This breaks the auto-correlation between the consecutive examples and makes the algorithm more stable and sample efficient. However, for non-stationary environments, past experiences might be outdated. For this reason, we don’t use a replay buffer. Instead, we break the auto-correlations by collecting experiences from many independent environments being sampled in parallel. After a batch of experiences is used we discard them. In our experiments, we simulated 400 environments in parallel and collected one experience step from each agent at each environment to form a training batch.

\subsection{Denser reward function}
\label{sec:denser_reward}
In some situations, we used a denser version of the evolutionary reward to speed up the training process. We call it the \textit{sugary} reward, $r_t^{\prime i} = \sum_{j \in \mathcal{A}_t}{k(\bm{g}^i, \bm{g}^j)}f_t^j$ where $f_t^j$ is the food collected by agent $j$ at the time instant $t$. In these simple environments, the \textit{sugary} and the evolutionary reward are almost equivalent since a family with more members will be able to collect more food and vice-versa. However, the \textit{sugary} reward contains more immediate information whilst the evolutionary reward has a lag between good (bad) actions and high (low) rewards; a family that is not doing a good job at collecting food will take a while to see some of its members die from starvation. Nonetheless, the evolutionary reward is more correct since it describes exactly what we want to maximise. Note that this reward was not used to produce the results when comparing E-VDN with CMA-ES.

\begin{figure}
  \centering
  \includegraphics[scale=0.4]{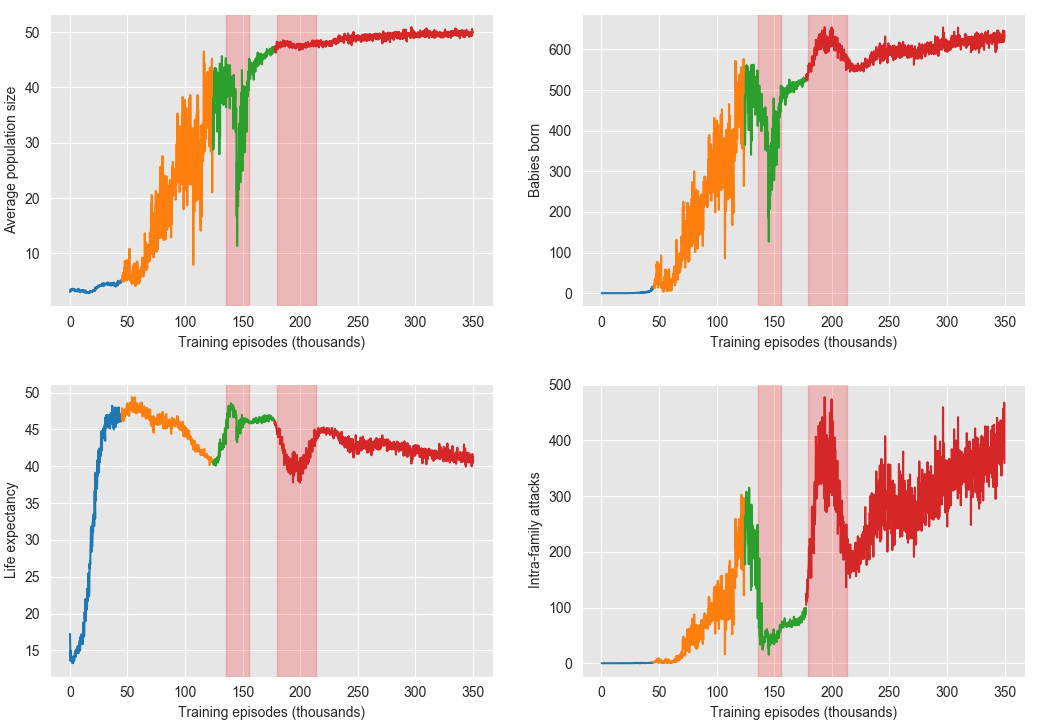}
  \caption{Macro-statistics when evolving bacteria using the standard evolutionary reward. I: learning to survive (blue line), II: learning to reproduce (orange), III: learning to detect kin (green), IV: learning to self-sacrifice (red). The red bands correspond to the First and Second Family Wars.}
  \label{fig:standard_history}
\end{figure}

When using the standard evolutionary reward to evolve the larger NNs, the same four eras, that were observed with the \textit{sugary} reward, emerge. However, their progression is not as linear. In this case, the families take longer to learn and sometimes one family evolves much faster than the others. When this happens, the families left behind eventually catch up with the most developed ones. The behaviour of the emerging families successfully interferes with the developed ones creating a temporary disruption in the environment which disrupts its macro-statistics. Two disruptions were observed in one of our simulations and we named them the First and the Second Family Wars (fig. \ref{fig:standard_history}).

\section{Results details}
\subsection{Cannibalism and suicide as a tool for gene survival}
\label{sec:app_can}
\begin{figure}
  \centering
  \includegraphics[scale=0.5]{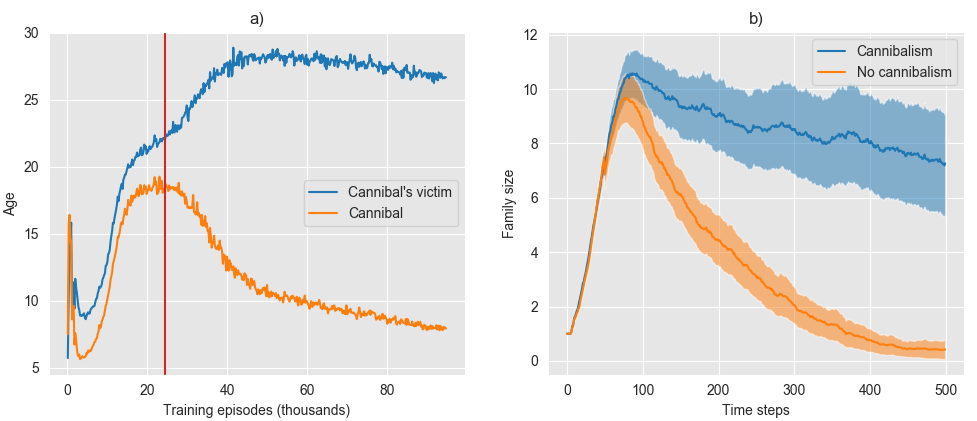}
  \caption{a) The average age of intra-family cannibals and cannibals’ victims. The vertical red line marks the start of era IV. b) the size of family 1 averaged along 90 test episodes. To compute the orange line we simply blocked all the attacks between members of the family 1. The shaded bands represent the 95\% confidence interval.
}
  \label{fig:cannibalism_age}
\end{figure}

In the evolutionary history of the asexual environment we saw the rise of cannibalism in the fourth era. Figure \ref{fig:cannibalism_age}.a shows how the average age of cannibals and their victims grows apart in this era. After observing this behaviour, we wanted to know how important cannibalism was for gene survival. To answer this question, we measured the family size of a certain family when its members were not allowed to attack each other and compared it with the normal situation where intra-family attacks were allowed (see figure \ref{fig:cannibalism_age}.b). Figure \ref{fig:cannibalism_age}.b clearly shows that, in this environment, cannibalism is essential for long-term gene survival. We also ran this exact experiment before the fourth era and achieved the opposite results, suggesting that before this era the agents didn't yet know how to use cannibalism to their gene's advantage (results not shown).

\subsection{Genetic drift}
The environment starts with a unique set of alleles\footnote{A gene can have many alleles — a variant form of the same gene. For example, a gene that codes the eye colour may have two different alleles one for brown eyes and other for blue eyes. In everyday language, the term gene is often used when referring to an allele, when one says that two siblings carry the same eye colour gene they actually mean that they carry the same allele. We will use this language as well, but it should be clear from the context to which one we are referring.}, and there is no mechanism to add diversity to this initial set (no mutations) but there is a mechanism to remove diversity: death. When a death occurs, there is a chance that the last carrier of a particular allele is lost and if there are no mutations, this is a permanent loss in diversity. This evolutionary mechanism that changes allele frequencies using chance events is called genetic drift. In the asexual environment, this means that if we simulate the environment for long enough all the agents will end up sharing the same allele. In the sexual environment, all the agents will have the same genome, however, this genome will likely be composed by alleles coming from all the five founders (see Video 2).

\begin{figure}
  \centering
  \includegraphics[scale=0.5]{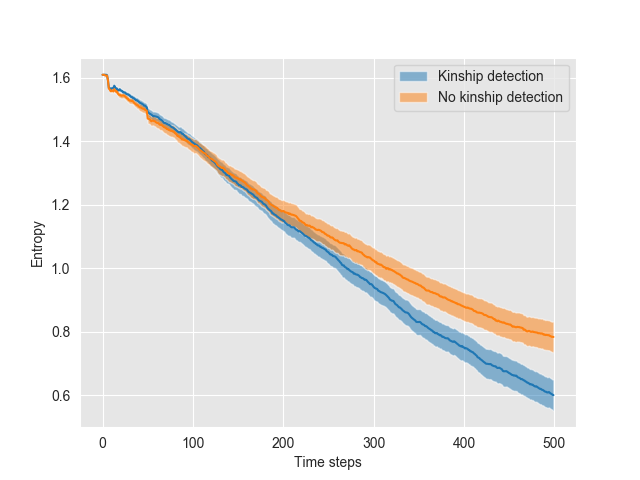}
  \caption{Entropy (diversity) in the allele frequency along the first 500 time steps in the situation where agents can detect kinship and when they can’t. The bands show the 95\% confidence interval.}
  \label{fig:genetic_drift}
\end{figure}

We found that in the asexual environment kin detection speeds up this decrease in allele diversity. This was expected since agents cooperate with kin and compete with non-kin. Therefore, as a family gets bigger, its members become more likely to encounter cooperative family members rather than competitive unrelated agents. This improves the survival and reproduction success of that family, making it even bigger. Figure \ref{fig:genetic_drift} shows the decrease in diversity with and without kin selection (we removed kin selection by zeroing out the kinship feature in the agents' observations). From the figure, it is evident that kin selection speeds up the decrease in diversity. However, note that before the 100\textsuperscript{th} iteration, kin detection leads to a slightly higher diversity. This happens because kin detection reduces intra-family violence, leading to fewer deaths and consequently to a slower genetic drift. The environment usually reaches its maximum capacity around the 100\textsuperscript{th} iteration, at this time the inter-family competition is at its highest and the positive feedback loop created by kin selection starts having a larger importance.

\end{appendices}

\end{document}